%% file: main.tex
\documentclass[10pt,journal,compsoc]{IEEEtran}
\usepackage{amsmath,amsfonts}
\usepackage{array}
\usepackage[caption=false,font=normalsize,labelfont=sf,textfont=sf]{subfig}
\usepackage{textcomp}
\usepackage{stfloats}
\usepackage{url}
\usepackage{verbatim}
\usepackage{graphicx}
\usepackage{cite}
\usepackage{multirow}
\usepackage{booktabs}

\usepackage{xspace}
\newcommand{\name}[0]{NAGphormer\xspace} 
\newcommand{\tname}[0]{NAGphormer${+}$\xspace}

\newcommand{\pname}[0]{Hop2Token\xspace}
\newcommand{\DAname}[0]{Neighborhood Augmentation\xspace}
\newcommand{\LNA}[0]{Local Neighborhood Augmentation\xspace}
\newcommand{\GNA}[0]{Global Neighborhood Augmentation\xspace}

\newcommand{\lna}[0]{LNA\xspace}
\newcommand{\gna}[0]{GNA\xspace}
\newcommand{\daname}[0]{NrAug\xspace}

\usepackage{bbding}
\makeatletter
\DeclareRobustCommand\onedot{\futurelet\@let@token\@onedot}
\def\@onedot{\ifx\@let@token.\else.\null\fi\xspace}
\def\eg{\emph{e.g}\onedot} 
\def\ie{\emph{i.e}\onedot}

\def\etal{\emph{et al}\onedot}

\usepackage{scalerel}
\usepackage{tikz}
\usetikzlibrary{svg.path}
\definecolor{orcidlogocol}{HTML}{A6CE39}
\tikzset{
    orcidlogo/.pic={
        \fill[orcidlogocol] svg{M256,128c0,70.7-57.3,128-128,128C57.3,256,0,198.7,0,128C0,57.3,57.3,0,128,0C198.7,0,256,57.3,256,128z};
        \fill[white] svg{M86.3,186.2H70.9V79.1h15.4v48.4V186.2z}
        svg{M108.9,79.1h41.6c39.6,0,57,28.3,57,53.6c0,27.5-21.5,53.6-56.8,53.6h-41.8V79.1z M124.3,172.4h24.5c34.9,0,42.9-26.5,42.9-39.7c0-21.5-13.7-39.7-43.7-39.7h-23.7V172.4z}
        svg{M88.7,56.8c0,5.5-4.5,10.1-10.1,10.1c-5.6,0-10.1-4.6-10.1-10.1c0-5.6,4.5-10.1,10.1-10.1C84.2,46.7,88.7,51.3,88.7,56.8z};
    }
}
\newcommand\orcidicon[1]{\href{https://orcid.org/#1}{\mbox{\scalerel*{
                \begin{tikzpicture}[yscale=-1,transform shape]
                \pic{orcidlogo};
                \end{tikzpicture}
            }{|}}}}
\usepackage[implicit=false]{hyperref}
\hypersetup{hidelinks,
	colorlinks=true,
	allcolors=black,
	pdfstartview=Fit,
	breaklinks=true}

\usepackage{algorithmicx}
\usepackage{algpseudocode}  
\usepackage{algorithm}  

\begin{document}

\title{Tokenized Graph Transformer with Neighborhood Augmentation for Node Classification in Large Graphs}

\author{Jinsong Chen, 
Chang Liu, 
Kaiyuan Gao,
Gaichao Li, 
Kun He\textsuperscript{\orcidicon{0000-0001-7627-4604}}, \IEEEmembership{Senior~Member,~IEEE}

\IEEEcompsocitemizethanks{\IEEEcompsocthanksitem J. Chen and G. Li are with Institute of Artificial Intelligence, Huazhong University of Science and Technology; 
School of Computer Science and Technology, Huazhong University of Science and Technology; 
and Hopcroft Center on Computing Science, Huazhong University of Science and Technology, Wuhan 430074,  China. 

\IEEEcompsocthanksitem C. Liu, K. Gao and K. He are with School of Computer Science and Technology, Huazhong University of Science and Technology; and Hopcroft Center on Computing Science, Huazhong University of Science and Technology, Wuhan 430074,  China. 

}

\thanks{Manuscript received April ??th, 2023; accepted ??, 202?. }
\thanks{This work was supported by National Natural Science Foundation (62076105,U22B2017).}
\thanks{The first two authors contribute equally.}
\thanks{(Corresponding author: Kun He. E-mail:brooklet60@hust.edu.cn.)}
}

\markboth{Journal of \LaTeX\ Class Files,~Vol.~14, No.~8, August~2021}%
{Chen \MakeLowercase{\textit{et al.}}: NAGphormer with Neighborhood Augmentation for Node Classification in Large Graphs}


\IEEEtitleabstractindextext{%
\begin{abstract}

Graph Transformers, emerging as a new architecture for graph representation learning,
suffer from the quadratic complexity on the number of nodes when handling large graphs.
To this end, we propose a \textbf{N}eighborhood \textbf{Ag}gregation Gra\textbf{ph} Transf\textbf{ormer} (\name) 
that treats each node as a sequence containing a series of tokens constructed by our proposed \pname module. 
For each node, \pname aggregates the neighborhood features from different hops into different representations, producing a sequence of token vectors as one input.  
In this way, \name could be trained in a mini-batch manner and thus could scale to large graphs.
Moreover, we mathematically show that compared to a category of advanced Graph Neural Networks (GNNs), called decoupled Graph Convolutional Networks, \name could learn more informative node representations from multi-hop neighborhoods.
In addition, we propose a new data augmentation method called  \textbf{N}eighbo\textbf{r}hood \textbf{Aug}mentation (\daname) based on the output of \pname that augments simultaneously the features of neighborhoods from global as well as local views to strengthen the training effect of \name.
Extensive experiments on benchmark datasets from small to large 
demonstrate the superiority of \name against existing graph Transformers and mainstream GNNs, and the effectiveness of \daname for further boosting \name.

\end{abstract}

}

\maketitle

\section{Introduction}
\input{0Intro}

\section{Related Work}
\input{1RW}

\section{Preliminaries}
\input{2BG}

\section{\name}
\input{3NAG}

\section{\DAname}
\input{5DA}

\section{Experiments}
\input{6Exp}

\section{Conclusion}
In this paper, we first propose \name, a novel and powerful graph Transformer for the node classification task.
Through two novel components, \pname and attention-based readout function, \name can handle large-scale graphs and adaptively learn the node representation from multi-hop neighborhoods.
The theoretical analysis indicates that \name can learn more expressive node representations than the decoupled GCN.
Based on the property of \pname, we further propose \daname, a novel data augmentation method augmenting the neighborhood information from global and local perspectives, to enhance the training effect of \name.

Experiments on various datasets from small to large demonstrate 
the superiority of \name over representative graph Transformers and Graph Neural Networks, and the effectiveness of proposed \daname in strengthening the performance of \name. 
Further ablation study shows that the effectiveness of structural encoding and attention-based readout function in \name, followed by the parameter studies. 
And analysis of the two components of \daname shows that both LNA and GNA are useful for boosting the model  performance. 
In the end, we show that \name and \tname are efficient on memory and running time. 
We can conclude that our tokenized design makes graph Transformers possible to handle large graphs.

\section*{Acknowledgments}
This work is supported by National Natural Science Foundation (U22B2017,62076105).

%

\bibliographystyle{IEEEtran}
\bibliography{reference}

\vspace{-0.6cm}
\begin{IEEEbiography}[{\includegraphics[width=1in,height=1.25in,clip,keepaspectratio]{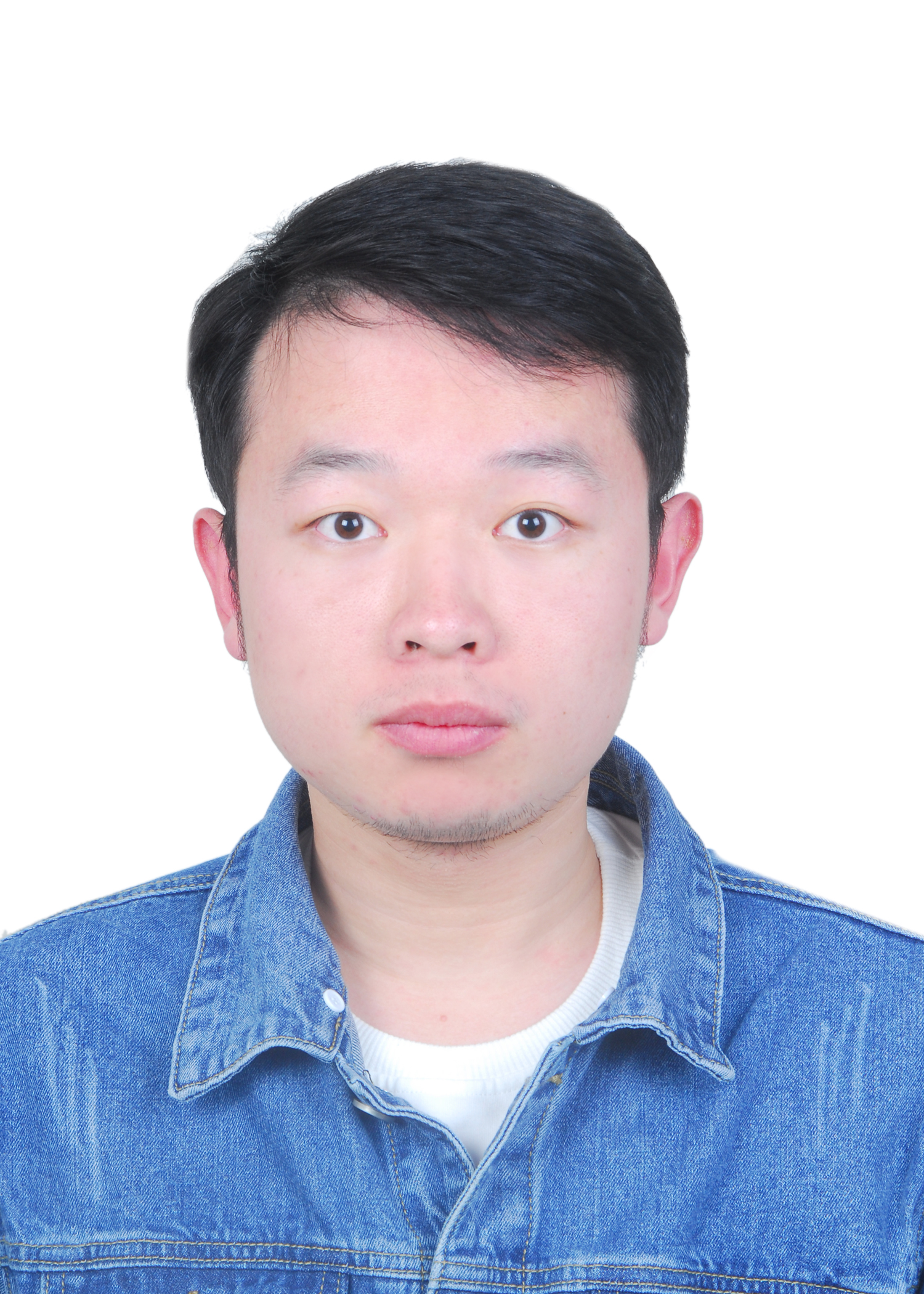}}]
{Jinsong Chen}
received his M.S.\ degree from Beijing University of Posts and Telecommunications in 2020. 
He is currently working toward the Ph.D. degree in School of Computer Science and Technology, Huazhong University of Science and Technology. 
His research interests include social network, graph embedding and graph neural networks.
\end{IEEEbiography}

\vspace{-0.6cm}
\begin{IEEEbiography}[{\includegraphics[width=1in,height=1.25in,clip,keepaspectratio]{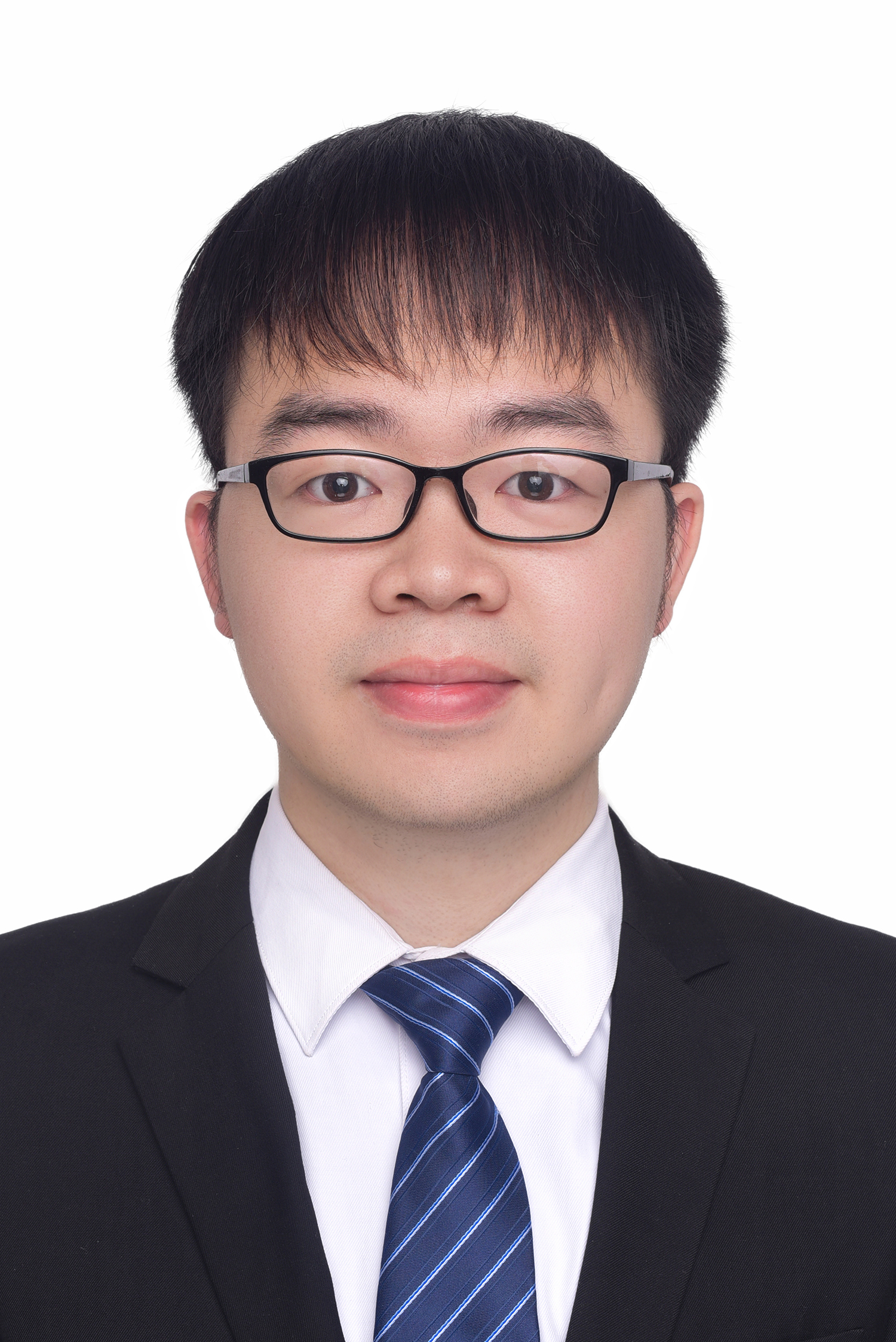}}]
{Chang Liu}
received his B.S.\ degree from Northeastern University, Shen Yang, China, in 2021. 
He is currently working toward the M.S.\ degree in School of Computer Science and Technology, Huazhong University of Science and Technology, Wuhan, China.
His research interests include graph neural networks, data augmentation and computer vision.
\end{IEEEbiography}

\vspace{-0.6cm}
\begin{IEEEbiography}[{\includegraphics[width=1in,height=1.25in,clip,keepaspectratio]{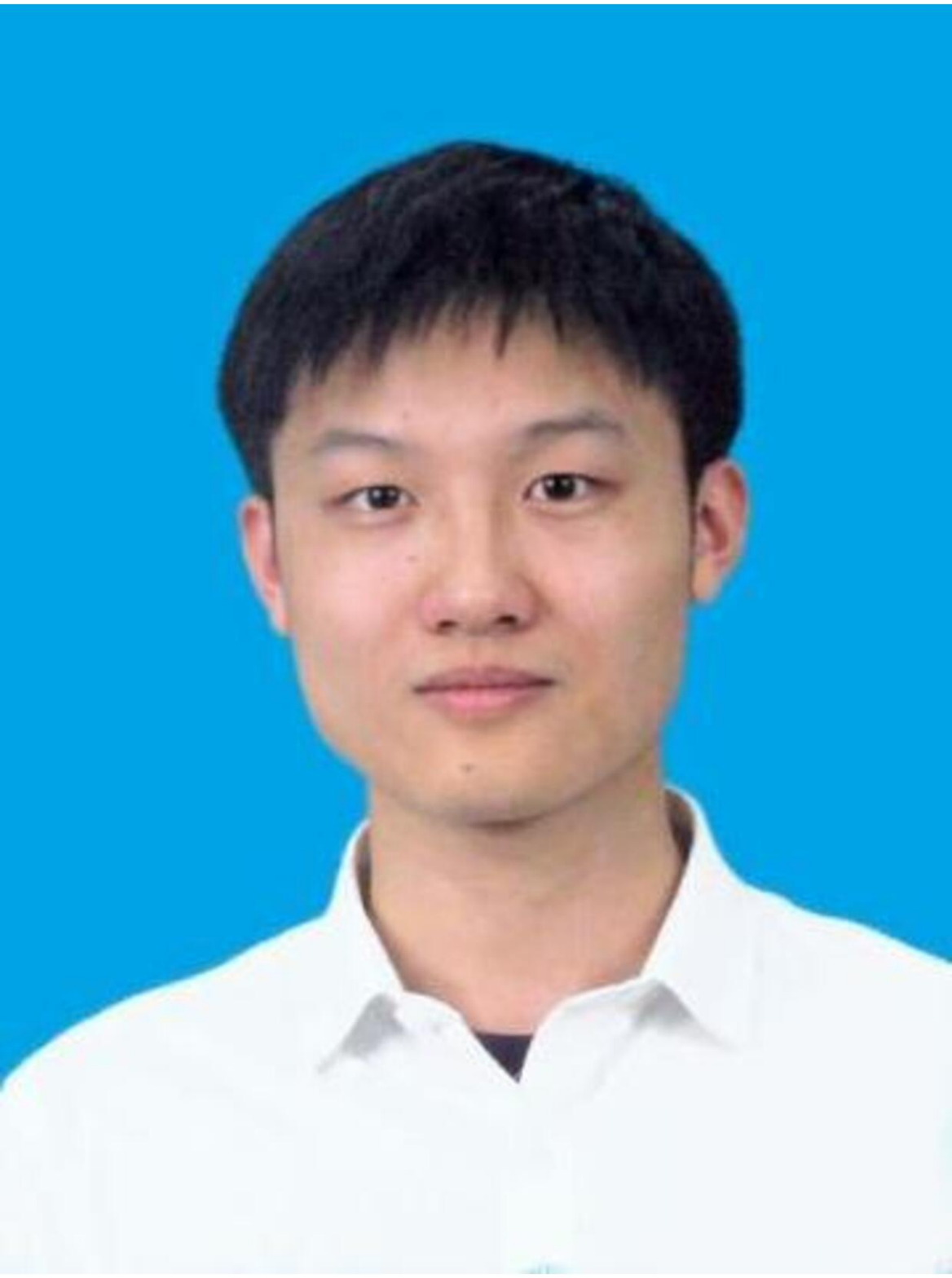}}]
{Kaiyuan Gao}
received his B.S. degree from Huazhong University of Science and Technology, Wuhan, China, in 2021. He is currently working toward the Ph.D. degree in School of Computer Science and Technology, Huazhong University of Science and Technology. His research interests include graph representation learning, molecule structure modeling.
\end{IEEEbiography}

\vspace{-0.6cm}
\begin{IEEEbiography}[{\includegraphics[width=1in,height=1.25in,clip,keepaspectratio]{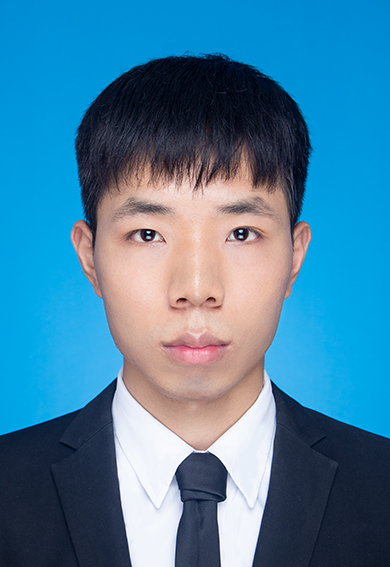}}]
{Gaichao Li}
received his M.S. degree from Wuhan University, Wuhan, China, in 2019. He is currently working toward the Ph.D. degree in School of Artificial Intelligence and Automation, Huazhong University of Science and Technology. His research interests include graph representation learning and social network.
\end{IEEEbiography}

\vspace{-0.6cm}
\begin{IEEEbiography}[{\includegraphics[width=1in,height=1.25in,clip,keepaspectratio]{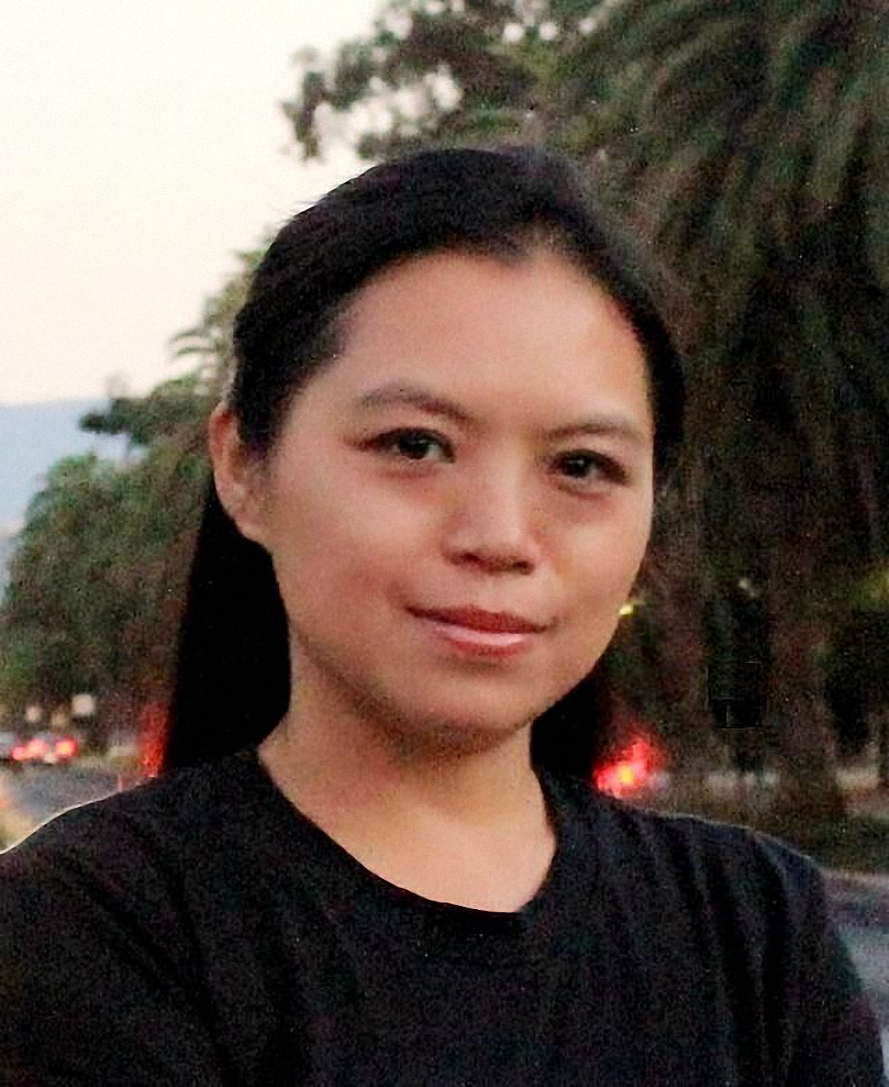}}]{Kun He} (SM18) received the Ph.D. degree in system engineering from Huazhong University of Science and Technology, Wuhan, China, in 2006. She is currently a Professor in School of Computer Science and Technology, Huazhong University of Science and Technology, Wuhan, China. She had been with the Department of Management Science and Engineering at Stanford University in 2011-2012 as a visiting researcher. 
She had been with the department of Computer Science at Cornell University in 2013-2015 as a visiting associate professor, in 2016 as a visiting professor, and in 2018 as a visiting professor.  She was honored as a Mary Shepard B. Upson visiting professor for the 2016-2017 Academic year in Engineering, Cornell University, New York. Her research interests include adversarial learning, representation learning, social network analysis, and combinatorial optimization.  
\end{IEEEbiography}

\vfill

\vfill

\end{document}

%% file: 0Intro.tex
\IEEEPARstart{G}{raphs}, as a powerful data structure, are widely used to represent entities and their relations in a variety of domains, such as social networks in sociology and protein-protein interaction networks in biology.
Their complex features (\eg, attribute features and topology features) make the graph mining tasks very challenging. 

Graph Neural Networks (GNNs)~\cite{GCNII,GCN,GAT}, owing to the message passing mechanism that aggregates neighborhood information for learning the node representations~\cite{mpnn}, have been recognized as a type of powerful deep learning techniques for graph mining tasks~\cite{GIN,gnnrs1,gnnrs2,gnnlink,cdgnn} over the last decade. 
Though effective, message passing-based GNNs have a number of inherent limitations, 
including over-smoothing~\cite{oversmoothing} and over-squashing~\cite{oversq} with the increment of model depth, limiting their potential capability for graph representation learning.
Though recent efforts~\cite{reoversm,skipnode,tackingos,pasl} have been devoted to alleviating the impact of 
these issues, the negative influence of their inherent limitations cannot be eliminated completely. 

Transformers~\cite{Transformer}, on the other hand recently, are well-known deep learning architectures that have shown superior performance in a variety of data with an underlying Euclidean or grid-like structure, such as natural languages~\cite{bert,liu2019roberta} and images~\cite{vit,swin}.
Due to their great modeling capability, there is a growing interest in generalizing Transformers to non-Euclidean data like graphs~\cite{GT,SAN,Graphormer,GraphTrans}. 
However, graph-structured data generally contain more complicated properties, including structural topology and attribute features, that cannot be directly encoded into Transformers as the tokens.

Existing graph Transformers have developed three techniques to address this 
challenge~\cite{trans-survey}: 
introducing structural encoding~\cite{GT,SAN}, 
using GNNs as auxiliary modules~\cite{GraphTrans}, 
and incorporating graph bias into the attention matrix~\cite{Graphormer}.
By integrating structural information into the model, 
graph Transformers exhibit competitive performance on various graph mining tasks, outperforming GNNs on node classification~\cite{SAN,sat} and graph classification~\cite{Graphormer,GraphTrans} tasks in small to mediate scale graphs.

Despite effectiveness, we observe that existing graph Transformers treat the nodes as independent tokens and construct a single sequence composed of all the node tokens to train Transformer model, 
leading to 
a quadratic complexity on the number of nodes for the self-attention calculation. 
Training such a model on large graphs will cost a huge amount of GPU resources that are generally unaffordable since the mini-batch training is unsuitable for graph Transformers using a single long sequence as the input.
Meanwhile, effective strategies that make GNNs scalable to large-scale graphs, including node sampling~\cite{FastGCN,LADIES} and approximation propagation~\cite{GBP,Grand+}, are not directly applicable to graph Transformers, as they capture the global attention of all node pairs and are independent of the message passing mechanism.

Recent works~\cite{gps,nodeformer} apply various efficient attention calculation techniques~\cite{performer,bigbird} into graph Transformers to achieve the linear computational complexity on the number of nodes and edges. 
Unfortunately, a graph could contain a great quantity of edges.
For instance, a common benchmark dataset of Reddit contains around 23K nodes and 11M edges, making it hard to directly train the linear graph Transformer on such graphs~\cite{performer,bigbird}.
In other words, the current paradigm of graph Transformers makes it intractable to generalize to large graphs.

To address the above challenge, 
we propose a \textbf{N}eighborhood \textbf{Ag}gregation Gra\textbf{ph} Transf\textbf{ormer} (\name)  for node classification in large graphs. 
Unlike existing graph Transformers that regard the nodes as independent tokens,
\name treats each node as a sequence and constructs tokens for each node by a novel neighborhood aggregation module called \pname. 
The key idea behind \pname is to aggregate neighborhood features from multiple hops 
and transform each hop into a representation, which could be regarded as a token.
\pname then constructs a sequence for each node based on the tokens in different hops to preserve the neighborhood information. 
The sequences are then fed into a Transformer-based module for learning the node representations.
By treating each node as a sequence of tokens, \name could be trained in a mini-batch manner and hence can handle large graphs even on limited GPU resources. 

Considering that the contributions of neighbors in different hops differ to the final node representation, \name further provides an attention-based readout function to learn the importance of each hop adaptively.
Moreover, we provide theoretical analysis on the relationship between \name and an advanced category of GNNs, the decoupled Graph Convolutional Network (GCN)~\cite{pt,appnp,SGC,gprgnn}. 
The analysis is from the perspective of self-attention mechanism and \pname, indicating that 
\name is capable of learning more informative node representations from 
the multi-hop neighborhoods. 

In this paper, we extend our conference version~\cite{nagformer} by proposing a novel data augmentation method to further enhance the performance of \name.
Data augmentation methods are known as effective techniques to improve the training effort.
Recent graph data augmentation methods~\cite{dropedge,verma2021graphmix,LAGNN} focus on modifying the information of nodes or edges by generating new node features or graph topology structures during the training stage, showing promising effectiveness for strengthening the model performance.
Nevertheless, most graph data augmentation methods focus on nodes or edges and are tailored to GNNs, which is unsuitable for \name, a Transformer method built on the features of multi-hop neighborhoods.

\begin{figure*}[t]
    \centering
	\includegraphics[width=15cm]{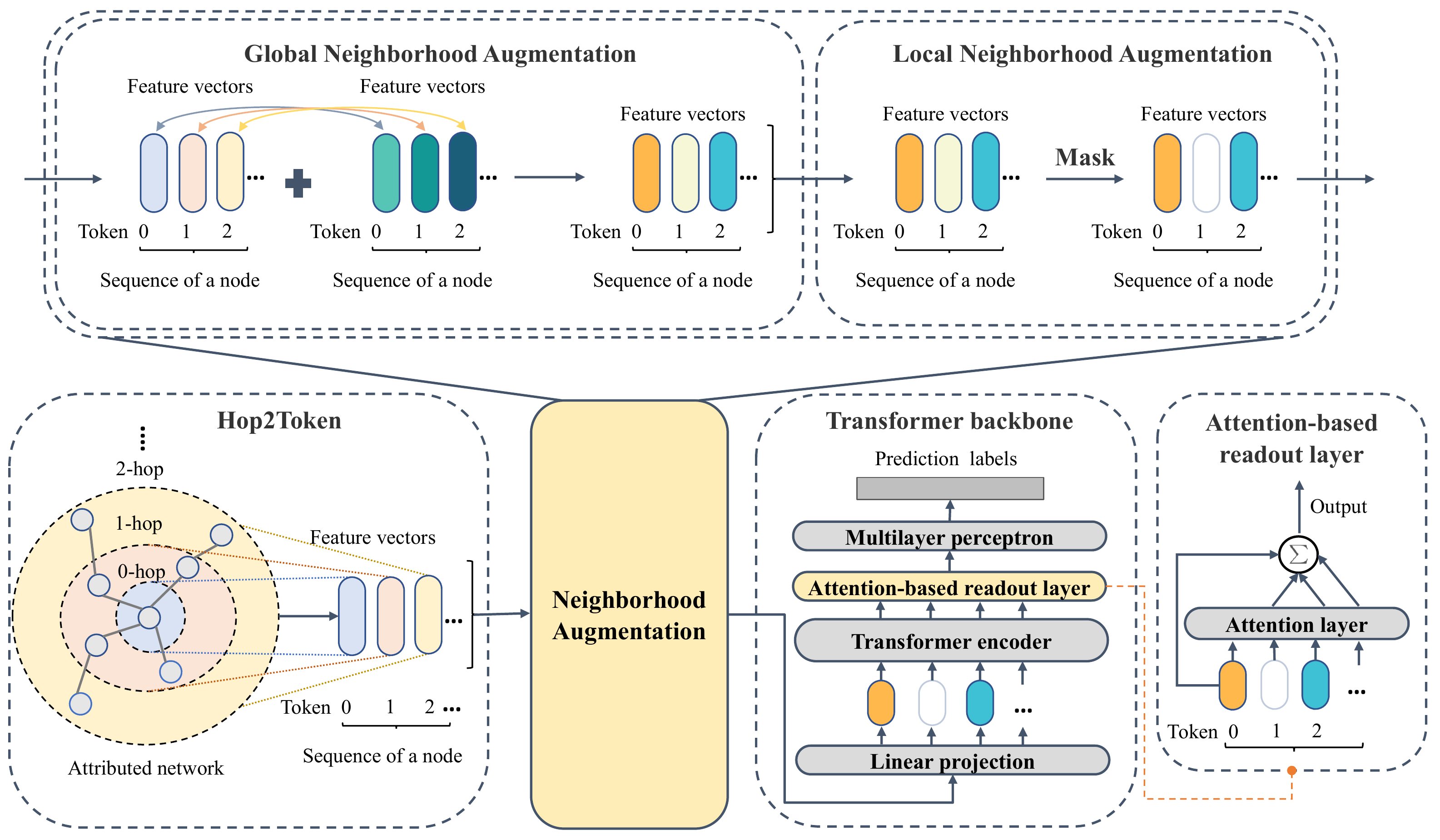}
	\caption{\textbf{The overall framework of \name with 
	Neighborhood Augmentation (\daname).}
 \name first uses a novel neighborhood aggregation module, \pname, to construct a sequence for each node based on the tokens of different hops of neighbors. 
 Then \daname is adopted to augment the information of multi-hop neighborhoods from both global and local perspectives.
 After the data augmentation process, \name learns the node representations using a Transformer backbone, and an attention-based readout function is developed to adaptively aggregate neighborhood information of different hops. 
 An MLP-based module is used in the end for label prediction.
	}
	\label{fig:fw}
\end{figure*}

Benefited from \pname that transforms the graph information of each node into the sequence of multi-hop neighborhoods, we introduce a new data augmentation method, \textbf{N}eighbo\textbf{r}hood \textbf{Aug}mentation (\daname), to augment the data obtained by \pname from the perspective of global mixing and local destruction. 
During the model training, \daname is applied to each sequence obtained from \pname with a fixed probability. 
First, we mix one sequence with another 
within the same batch and interpolating their labels accordingly. 
Then \daname masks a portion of the sequence to get the data for subsequent network. 
The advantage of this method is that it can fully utilize the neighborhood information of multiple nodes and destroy the data appropriately to reduce the risk of overfitting.
The overall framework is shown in Figure \autoref{fig:fw}.

We conduct extensive experiments on various popular benchmarks, including six small datasets and three large datasets, and the results demonstrate the superiority of the proposed method. 
The main contributions of this work are as follows:

\begin{itemize}
\item We propose \pname, a novel neighborhood aggregation method that aggregates the neighborhood features from each hop into a node representation, resulting in a sequence of token vectors that preserves neighborhood information for different hops. In this way, we can regard each node in the complex graph data as a sequence of tokens, and treat them analogously as in natural language processing and computer vision fields.

\item We propose a new graph Transformer model, \name, for the node classification task. \name can be trained in a mini-batch manner depending on the output of \pname, and therefore enables the model to handle large graphs. 
We also develop an attention-based readout function to adaptively learn the importance of different-hop neighborhoods to boost the model performance.

\item We prove that from the perspective of the self-attention mechanism, compared to an advanced category of GNNs, the decoupled GCN, the proposed \name can learn more expressive node representations from the multi-hop neighborhoods. 

\item We further propose a novel data augmentation method \daname that augments the neighborhood information obtained by \pname from both global and local perspectives to enhance the training effect of \name.

\item Extensive experiments on benchmark datasets from small to large demonstrate that \name consistently outperforms existing graph Transformers and mainstream GNNs. 
And the proposed \daname can further boost the performance of \name effectively.
\end{itemize}

%% file: 1RW.tex
\subsection{Graph Neural Network}
Graph Neural Network (GNN) has become a powerful technique for modeling graph-structured data.
Based on the message-passing mechanism, GNN can simultaneously learn the node representations from topology features and attribute features.
Typical GNNs, such as GCN~\cite{GCN} and GAT~\cite{GAT}, leverage the features of immediate neighbors via different aggregation strategies to learn the node representations, exhibiting competitive performance on various graph mining tasks.
However, typical GNNs obey the coupled design that binds the aggregation and feature transformation modules in each GNN layer, leading to the over-smoothing~\cite{oversmoothing} and over-squashing issues~\cite{oversq} on deep-layer GNNs. 
Such a problem limits the model's ability to capture deep graph structural information.

A reasonable solution is to decouple the aggregation and feature transformation modules in each GNN layer, treating them as independent modules~\cite{appnp,SGC,gprgnn}, termed decoupled Graph Convolutional Network (decoupled GCN)~\cite{pt}.
Decoupled GCN utilizes various propagation methods, such as personalized PageRank~\cite{appnp} and random walk~\cite{SGC}, to aggregate features of multi-hop neighborhoods and further generate the node representations.
Since the nonlinear activation functions between GNN layers are removed, decoupled GCN exhibits high computational efficiency and has become an advanced type of GNNs in recent years.

Besides the decoupled strategy, recent works~\cite{reoversm,skipnode,tackingos,pasl} make efforts to address the over-smoothing and over-squashing issues by developing novel training tricks~\cite{reoversm,tackingos} or new graph neural network architectures~\cite{skipnode, pasl}.
By introducing carefully designed techniques, the impact of over-smoothing and over-squashing problems in GNNs could be well alleviated.

Most GNNs~\cite{GCNII,SimpGNN,GCN,GAT} require the entire adjacency matrix as the input during training. 
In this way, when applying to large-scale graphs, the cost of training is too high to afford. 
There are two categories of strategies for generalizing GNN to large-scale graphs:

(\uppercase\expandafter{\romannumeral1}) 
The node sampling strategy~\cite{GraphSAGE,ClusterGCN,LADIES,GraphSAINT} that samples partial nodes from the whole graph via different methods, such as random sampling from neighbors~\cite{GraphSAGE} and sampling from GNN layers~\cite{LADIES}, to reduce the size of nodes for model training.

(\uppercase\expandafter{\romannumeral2}) 
The approximation propagation~\cite{GBP,Grand+,PPRGo} that accelerates the propagation operation via several approximation methods, such as approximate PageRank~\cite{PPRGo} and sub-matrix approximation~\cite{Grand+}.

However, by designing various sampling-based or approximation-based methods to reduce the training cost, these models will inevitably lead to information loss and somehow restrict their performance on large-scale networks. 

\subsection{Graph Transformer}
In existing graph Transformers, there are three main strategies to incorporate graph structural information into the Transformer architecture so as to learn the node representations:

(\uppercase\expandafter{\romannumeral1}) 
Extracting the positional embedding from graph structure. 
Dwivedi~\etal~\cite{GT} utilize Laplacian eigenvectors to represent positional encodings of the original Transformer and fuse them with the raw attributes of nodes as the input. Derived from~\cite{GT}, Devin~\etal~\cite{SAN} leverage the full spectrum of Laplacian matrix to learn the positional encodings.

(\uppercase\expandafter{\romannumeral2})
Combining GNN and Transformer. 
In addition to representing structural information by the eigenvectors, Wu~\etal~\cite{GraphTrans} regard GNNs as an auxiliary module to extract fixed local structural information of nodes and further feed them into the Transformer to learn long-range pairwise relationships.
Chen~\etal~\cite{sat} utilize a GNN model as the structure extractor to learn different types of structural information, such as $k-$subtree and $k-$subgraph, to capture the structure similarity of node pairs via the self-attention mechanism.
Rampášek~\etal~\cite{gps} develop a hybrid layer that contains a GNN layer and a self-attention layer to capture both local and global information.

(\uppercase\expandafter{\romannumeral3})
Integrating the graph structural bias into the self-attention matrix.
There are several efforts to transform various graph structure features into attention biases and integrate them into the self-attention matrix to enable the Transformer to capture graph structural information. Ying~\etal~\cite{Graphormer} propose a spatial encoding method that models the structural similarity of node pairs based on the length of their shortest path. 
Zhao~\etal~\cite{gophormer} propose a proximity-enhanced attention matrix by considering the relationship of node pairs in different neighborhoods. 
Besides, by modeling edge features in chemical and molecular graphs, Dwivedi~\etal~\cite{GT} extend graph Transformers to edge feature representation by injecting them into the self-attention module of Transformers.
Hussain~\etal~\cite{egt} utilize the edge features to strengthen the expressiveness of the attention matrix.
Wu~\etal~\cite{nodeformer} introduce the topology structural information as the relational bias to strengthen the original attention matrix. 

Nevertheless, the computational complexity of most existing graph Transformers is quadratic with the number of nodes.
Although GraphGPS~\cite{gps} and NodeFormer~\cite{nodeformer} achieve linear complexity with the number of nodes and edges by introducing various linear Transformer backbones, 
Such high complexity makes these methods hard to directly handle graph mining tasks on large-scale networks with millions of nodes and edges since they require the entire graph as the input.

Recent works~\cite{gophormer,ansgt} sample several ego-graphs of each node and then utilize Transformer to learn the node representations on these ego-graphs so as to reduce the computational cost of model training.
However, the sampling process is still time-consuming in large graphs.
Moreover, the sampled ego-graphs only contain limited neighborhood information due to the fixed and small sampled graph size for all nodes, which is insufficient to learn the informative node representations.

\subsection{Graph Data Augmentation}
\label{Graph Data Augmentation}

Most current data augmentation techniques involve modifying existing data directly or generating new data with the same distribution using existing training data. 
However, graph data are irregular and non-Euclidean structures, making developing data augmentation techniques for graphs challenging. 
Existing graph data augmentation methods can be categorized into three groups: node augmentation, edge augmentation, and feature augmentation.

Node augmentation methods attempt to operate on nodes in the graph. Wang~\etal~\cite{wang2021mixup} propose a method that interpolates a pair of nodes and their ground-truth labels to produce a novel and synthetic sample for training. Verma~\etal~\cite{verma2021graphmix} present GraphMix, which trained an auxiliary Fully-Connected Network to generate better features using the node features. 
Feng~\etal~\cite{dropnode} propose DropNode, which removes the entire feature vector for some nodes to enhance the model robustness.

Edge augmentation methods modify the graph connectivity by adding or removing edges. The most representative work is DropEdge~\cite{dropedge}, which randomly removes some edges from the input graph and can be plugged into exiting popular GCNs to improve the performance. Another approach is to update the graph structure with the model's predicted results, such as AdaEdge~\cite{AdaEdge}, GAUG~\cite{GAUG}, and MH-Aug~\cite{MH-Aug}.

Feature augmentation methods seek to augment node features for better performance. FLAG~\cite{FLAG} improves the generalization ability of GNNs through gradient-based adversarial perturbation. 
LAGNN~\cite{LAGNN} learns the distribution of the neighbor's node representation based on the central node representation and uses the resulting features with the raw node features to enhance the representation of GNN.

Unlike the ideas of previous studies, which augment graph data from the perspective of nodes or edges, we propose a new augmentation method based on the output of \pname and augments graph data from the perspective of neighborhood information.

%% file: 2BG.tex
\subsection{Problem Formulation}
Let $G=(V,E)$ be an unweighted and undirected attributed graph, where $V=\{v_1, v_2, \cdots, v_n\}$, and $n=|V|$. 
Each node $v \in V$ has a feature vector $\mathbf{x}_v \in \mathbf{X}$, where $\mathbf{X} \in \mathbb{R}^{n \times d}$ is the feature matrix describing the attribute information of nodes and $d$ is the dimension of feature vector.
$\mathbf{A}\in \mathbb{R}^{n \times n}$ represents the adjacency matrix and  $\mathbf{D}$ is the diagonal degree matrix.
The normalized adjacency matrix is defined as
$\hat{\mathbf{A}}=\tilde{\mathbf{D}}^{-1 / 2} \tilde{\mathbf{A}} \tilde{\mathbf{D}}^{-1 / 2}$, where $\tilde{\mathbf{A}}$ denotes the adjacency matrix with self-loops and $\tilde{\mathbf{D}}$ denotes the corresponding diagonal degree matrix. 
The node classification task provides a labeled node set $V_l$ and an unlabeled node set $V_u$. Let $\mathbf{Y}\in \mathbb{R}^{n \times c}$ denote the label matrix where $c$ is the number of classes. Given the labels $\mathbf{Y}_{V_l}$, the goal is to predict the labels $\mathbf{Y}_{V_u}$ for unlabeled nodes.

\subsection{Graph Neural Network}
Graph Neural Network (GNN) has become a powerful technique to model the graph-structured data. 
Graph Convolutional Network (GCN)~\cite{GCN} is a typical model of GNN that applies the first-order approximation of spectral convolution~\cite{sp-gcn} to aggregate information of immediate neighbors.
A GCN layer can be written as:
\begin{equation}
	\mathbf{H}^{(l+1)} = \sigma(\hat{\mathbf{A}} \mathbf{H}^{(l)} \mathbf{W}^{(l)}),
	\label{eq:gcn}
\end{equation}
where $\mathbf{H}^{(l)} \in \mathbb{R}^{n\times d^{(l)}}$ and
$\mathbf{W}^{(l)} \in \mathbb{R}^{d^{(l)} \times d^{(l+1)}}$
denote the representation of nodes and the learnable parameter matrix in the $l$-th layer, respectively.
$\sigma(\cdot)$ denotes the non-linear activation function.

Eq. (\ref{eq:gcn}) contains two operations, \ie, neighborhood aggregation and feature transformation, which are coupled in the GCN layer. 
Such a coupled design would lead to the over-smoothing problem~\cite{oversmoothing} when the number of layers increases, limiting the model to capture deep structural information.
To address this issue, 
the decoupled GCN~\cite{appnp,SGC} separates the feature transformation and neighborhood aggregation in the GCN layer and treats them as independent modules.
A general form of decoupled GCN is described as~\cite{gprgnn}:
\begin{equation}
	\mathbf{Z}=\sum_{k=0}^{K} \beta_{k} \mathbf{H}^{(k)}, \mathbf{H}^{(k)}=\hat{\mathbf{A}} \mathbf{H}^{(k-1)}, \mathbf{H}^{(0)}=\boldsymbol{f}_{\theta}(\mathbf{X}),
	\label{eq:dgcn}
\end{equation}
where $\mathbf{Z}$ denotes the final representations of nodes, $\mathbf{H}^{(k)}$ denotes the hidden representations of nodes at propagation step $k$, $\beta_{k}$ denotes the aggregation coefficient of propagation step $k$,
$\hat{\mathbf{A}}$ denotes the normalized adjacency matrix,
$\boldsymbol{f}_{\theta}$ denotes a neural network module and $\mathbf{X}$ denotes the raw attribute feature matrix.
Such a decoupled design exhibits high computational efficiency and enables the model to capture deeper structural information. 

\subsection{Transformer} \label{trans-att}
The Transformer encoder~\cite{Transformer} contains a sequence of Transformer layers, where each layer is comprised of a multi-head self-attention (MSA) and a position-wise feed-forward network (FFN). 
The MSA module is the critical component that aims to capture the semantic correlation between the input tokens. For simplicity, we use the single-head self-attention module for description. 
Suppose we have an input $\mathbf{H}\in \mathbb{R}^{n \times d}$ for the self-attention module where $n$ is the number of tokens and $d$ the hidden dimension. The self-attention module first projects $\mathbf{H}$ into three subspaces, namely $\mathbf{Q}$, $\mathbf{K}$ and $\mathbf{V}$:
\begin{equation}
	\mathbf{Q} = \mathbf{H}\mathbf{W}^Q,~
	\mathbf{K} = \mathbf{H}\mathbf{W}^K,~
	\mathbf{V} = \mathbf{H}\mathbf{W}^V,
	\label{eq:QKV} 
\end{equation}
where $\mathbf{W}^Q\in \mathbb{R}^{d \times d_K}, \mathbf{W}^K\in \mathbb{R}^{d \times d_K}$ and $\mathbf{W}^V \in \mathbb{R}^{d \times d_V}$ are the projection matrices. 
The output matrix is calculated by:
\begin{equation}
	\mathbf{H}^{\prime} = \mathrm{softmax}\left(\frac{\mathbf{Q}\mathbf{K}^{\top}}{\sqrt{d_K}}\right)\mathbf{V}.
	\label{eq:att-out}
\end{equation}
The attention matrix, $\mathrm{softmax}\left(\frac{\mathbf{Q}\mathbf{K}^{\top}}{\sqrt{d_K}}\right)$, captures the pair-wise similarity of input tokens in the sequence. Specifically, it calculates the dot product between each token pair after projection. The softmax is applied row-wise.

%% file: 3NAG.tex
In this section, we present the proposed \name in detail. 
To handle graphs at scale, 
we first introduce a novel neighborhood aggregation module called \pname, 
then we build \name together with structural encoding and attention-based readout function.
We also provide the computational complexity of \name.
Finally, we conduct the theoretical analysis of \name, which brings deeper insights into the relation between \name and decoupled GCN.

\subsection{\pname}

How to aggregate information from adjacent nodes into the node representation is crucial in reasonably powerful Graph Neural Network (GNN) architectures. 
To inherit the desirable properties, we design \pname that considers the neighborhood information of different hops.

Speciffically, for each node $v$, let $\mathcal{N}^{k}(v) = \{u \in V | d(v, u)\leq k\}$ denote its $k$-hop neighborhood, where $d(v, u)$ represents distance of the shortest path between $v$ and $u$. We define $\mathcal{N}^{0}(v) = \{v\}$, \ie, the $0$-hop neighborhood is the node itself. 
In \pname, we transform the $k$-hop neighborhood $\mathcal{N}^{k}(v)$ into a neighborhood embedding $\mathbf{x}^k_v$ with an aggregation operator $\phi$. 
In this way, the $k$-hop representation of a node $v$ can be expressed as:
\begin{equation}
	\mathbf{x}^k_v = \phi(\mathcal{N}^{k}(v)).
	\label{eq:token}
\end{equation}
By Eq. (\ref{eq:token}), we can calculate the neighborhood embeddings for variable hops of a node and further construct a sequence to represent its neighborhood information, \ie,
$\mathcal{S}_{v} = (\mathbf{x}^0_v, \mathbf{x}^1_v,...,\mathbf{x}^K_v )$, where $K$ is fixed as a hyper-parameter.  
Assume $\mathbf{x}^k_v$ is a $d$-dimensional vector, the sequences of all nodes in graph $G$ will construct a tensor $\mathbf{X}_G \in \mathbb{R}^{n\times (K+1)\times d}$. 
To better illustrate the implementation of \pname, we decompose $\mathbf{X}_{G}$ to a sequence $\mathcal{S} = (\mathbf{X}_0, \mathbf{X}_1, \cdots, \mathbf{X}_K)$, where $\mathbf{X}_k \in \mathbb{R}^{n\times d}$ can be seen as the $k$-hop neighborhood matrix. Here we define $\mathbf{X}_0$ as the original feature matrix $\mathbf{X}$.

In practice, we apply a propagation process similar to the method in \cite{gprgnn,QT} to obtain the sequence of $K$-hop neighborhood matrices. 
Given the normalized adjacency matrix $\hat{\mathbf{A}}$ (\textit{aka} the transition matrix~\cite{diffusion}) and $\mathbf{X}$, multiplying $\hat{\mathbf{A}}$ with $\mathbf{X}$ aggregates immediate neighborhood information. Applying this multiplication consecutively allows us to propagate information at larger distances. For example, we can access the $2$-hop neighborhood information by $\hat{\mathbf{A}}(\hat{\mathbf{A}}\mathbf{X})$.
Thereafter, the $k$-hop neighborhood matrix can be described as:
\begin{equation}
	\mathbf{X}_{k} = \hat{\mathbf{A}}^{k}\mathbf{X}.
	\label{eq:fp}
\end{equation}

The detailed implementation is drawn in Algorithm~\ref{alg:h2t}.
\begin{algorithm}[t]  
	\caption{The \pname Algorithm}  
	\label{alg:h2t}  
	\begin{algorithmic}[1]  
		\Require
		Normalized adjacency matrix $\hat{\mathbf{A}}$;
		Feature matrix $\mathbf{X}$;
		Propagation step $K$
		\Ensure  
		Sequences of all nodes $\mathbf{X}_{G}$
	
		\For{$k=0$ to $K$}
		\For{$i=0$ to $n$}
		\State $\mathbf{X}_{G}[i,k]=\mathbf{X}[i]$;
        \EndFor
        \State
		$\mathbf{X} = \hat{\mathbf{A}}\mathbf{X}$;
		\EndFor 
         \\  
		\Return 		Sequences of all nodes $\mathbf{X}_{G}$;  
	\end{algorithmic}  
\end{algorithm}   
The advantages of \pname are two-fold.  
(\uppercase\expandafter{\romannumeral1}) \pname is a non-parametric method. It can be conducted offline before the model training, and the output of \pname supports mini-batch training. In this way, the model can handle graphs of arbitrary sizes, thus allowing the generalization of graph Transformer to large-scale graphs.
(\uppercase\expandafter{\romannumeral2}) Encoding $k$-hop neighborhood of a node into one representation is helpful for capturing the hop-wise semantic correlation, which is ignored in typical GNNs~\cite{GCN,appnp,gprgnn}.

\subsection{\name for Node Classification}

Given an attributed graph, besides the attribute information of nodes, the structural information of nodes is also a crucial feature for graph mining tasks.
Hence, we construct a hybrid feature matrix by concatenating the structural feature matrix to the attribute feature matrix to preserve the structural information and attribute information of nodes simultaneously.
Specifically,  We adopt the eigenvectors of the graph's Laplacian matrix to capture the nodes' structural information.
In practice, we select the eigenvectors corresponding to the $s$ smallest non-trivial eigenvalues to construct the structure matrix $\mathbf{U} \in \mathbb{R} ^ {n \times s}$~\cite{GT,SAN}. 
Then we combine the original feature matrix $\mathbf{X}$ with the structure matrix $\mathbf{U}$ to preserve both the attribute and structural information:
\begin{equation}
	\mathbf{X}^{\prime} = 
	\mathbf{X} \Vert \mathbf{U}.
	\label{eq:infea}
\end{equation}
Here $\Vert$ indicates the concatenation operator and $\mathbf{X}^{\prime} \in \mathbb{R} ^ {n \times (d+s)}$ denotes the fused feature matrix, which is then used as the input of \pname for calculating the information of different-hop neighborhoods.
Accordingly, the effective feature vector for node $v$ is extended as $\mathbf{x}^{\prime}_v \in \mathbb{R}^{1\times(d+s)}$.  

Next, we assemble an aggregated neighborhood sequence as $\mathcal{S}_{v} = (\mathbf{x}^{\prime 0}_v, \mathbf{x}^{\prime 1}_v,...,\mathbf{x}^{\prime K}_v$) by applying \pname. 
Then we map $\mathcal{S}_{v}$ to the hidden dimension $d_m$ of the Transformer with a learnable linear projection:
\begin{equation}
	\mathbf{Z}^{(0)}_{v}=\left[\mathbf{x}^{\prime 0}_v \mathbf{E};~ \mathbf{x}^{\prime 1}_v \mathbf{E};~ \cdots;~ \mathbf{x}^{\prime K}_v \mathbf{E}\right],
	\label{eq:ly_att}
\end{equation}
where $\mathbf{E} \in \mathbb{R}^{(d+s) \times d_m}$ and $\mathbf{Z}^{(0)}_{v} \in \mathbb{R}^{(K+1) \times d_m}$.

Then, we feed the projected sequence into the Transformer encoder. 
The building blocks of the Transformer contain multi-head self-attention (MSA) and position-wise feed-forward network (FFN). We follow the implementation of the vanilla Transformer encoder described in \cite{Transformer}, while LayerNorm (LN) is applied before each block~\cite{prenorm}. And the FFN consists of two linear layers with a GELU non-linearity:
\begin{align}
\mathbf{Z}^{\prime(\ell)}_{v} &=\operatorname{MSA}\left(\operatorname{LN}\left(\mathbf{Z}^{(\ell-1)}_{v}\right)\right)+\mathbf{Z}^{(\ell-1)}_{v}, \\
\mathbf{Z}^{(\ell)}_{v} &=\operatorname{FFN}\left(\operatorname{LN}\left(\mathbf{Z}^{\prime(\ell)}_{v}\right)\right)+\mathbf{Z}^{\prime(\ell)}_{v}, 
\end{align}
where $\ell=1, \ldots, L$ implies the $\ell$-th layer of the Transformer.

In the end, a novel readout function is applied to the output of the Transformer encoder. 
Through several Transformer layers, the corresponding output $\mathbf{Z}^{(\ell)}_{v}$ contains the embeddings for all neighborhoods of node $v$.
It requires a readout function to aggregate the information of different neighborhoods into one embedding. 
Common readout functions include summation and mean~\cite{GraphSAGE}. However, these methods ignore the importance of different neighborhoods. 
Inspired by GAT~\cite{GAT}, we propose an attention-based readout function to learn such importance by computing the attention coefficients between $0$-hop neighborhood (\ie, the node itself) and every other neighborhood.
Specifically, for the output matrix $\mathbf{Z}^{v}\in \mathbb{R}^{(K+1) \times d_{m}}$ of node $v$, $\mathbf{Z}^{v}_0$ is the token representation of the node itself and $\mathbf{Z}^{v}_k$ is its $k$-hop representation.  
We calculate the normalized attention coefficients for its $k$-hop neighborhood:
\begin{equation}
	\alpha^{v}_k = \frac{exp((\mathbf{Z}^{v}_0\Vert\mathbf{Z}^{v}_k)\mathbf{W}_{a}^{\top})}{\sum_{i=1}^{K}exp((\mathbf{Z}^{v}_0\Vert\mathbf{Z}^{v}_i)\mathbf{W}_{a}^{\top})} ,
	\label{eq:att_readout}
\end{equation}
where $\mathbf{W}_{a}\in \mathbb{R}^{1 \times 2d_{m}}$ denotes the learnable projection and $i=1, \ldots, K$. Therefore, the readout function takes the correlation between each neighborhood and the node representation into account.
The node representation is finally aggregated as follows:
\begin{equation}
	\mathbf{Z}^{v}_{out} = \mathbf{Z}^{v}_0 + \sum_{k=1}^{K} \alpha^{v}_k \mathbf{Z}^{v}_k.
	\label{eq:node_final}
\end{equation}

Based on Eq. (\ref{eq:node_final}), we could obtain the final representation matrix of all nodes $\mathbf{Z}_{out} \in \mathbb{R}^{n \times d_{m}}$.
We further utilize the Multilayer Perceptron (MLP) as the classifier to predict the labels of nodes:
\begin{equation}
	\hat{\mathbf{Y}} = \mathrm{MLP}(\mathbf{Z}_{out}),
	\label{eq:pre_label}
\end{equation}
where $\hat{\mathbf{Y}} \in \mathbb{R}^{n \times c}$ denotes the predicted label matrix of nodes.
And the loss function is described as follows:
\begin{equation}
        \mathcal{L} = -\sum_{i\in V_{l}}\sum_{j = 0}^{c} {\mathbf{Y}_{i,j}}\mathrm{ln}\hat{\mathbf{Y}}_{i,j}.
	\label{eq:loss}
\end{equation}           

\subsection{Computational Complexity Analysis}
We provide the computational complexity analysis of \name on time and space.

\textbf{Time complexity.} 
The time complexity of \name mainly depends on the self-attention module of the Transformer. So the computational complexity of \name is $O(n(K+1)^2d)$, where $n$ denotes the number of nodes, $K$ denotes the number of hops and $d$ is the dimension of parameter matrix (\ie, feature vector).

\textbf{Space complexity.} 
The space complexity is based on the number of model parameters and the outputs of each layer. The first part is mainly on the Transformer layer $O(d^2L)$, where $L$ is the number of Transformer layers. 
The second part is on the attention matrix and the hidden node representations, $O(b(K+1)^2 + b(K+1)d)$, where $b$ denotes the batch size. 
Thus, the total space complexity is $O(b(K+1)^2 + b(K+1)d+ d^2L)$.

The computational complexity analysis reveals that the memory cost of training \name on GPU devices is restricted to the batch size $b$.
Hence, with a suitable $b$, \name can handle graph learning tasks in large-scale graphs even on limited GPU resources.

\subsection{Theoretical analysis of \name}
In this subsection, we discuss the relation of \name and decoupled GCN 
through the lens of node representations of \pname and self-attention mechanism. We theoretically show that \name could learn more informative node representations from the multi-hop neighborhoods than decoupled GCN does.

\textbf{Fact 1.} \textit{From the perspective of the output node representations of \pname,
we can regard the decoupled GCN as applying a self-attention mechanism with a fixed attention matrix $\mathbf{S}\in \mathbb{R}^{(K+1)\times (K+1)}$, where $\mathbf{S}_{K,k}=\beta_{k}$ ($k\in\{0,...,K\}$) and other elements are all zeroes.  
}

Here $K$ denotes the total propagation step, $k$ represents the current propagation step,
$\beta_{k}$ represents the aggregation weight at propagation step $k$ in the decoupled GCN.  

\textit{Proof.} First, both \pname and decouple GCN utilize the same propagation process to obtain the information of different-hop neighborhoods. So we use the same symbol $\mathbf{H}_{i}^{(k)} \in \mathbb{R}^{1\times d}$ to represent the neighborhood information of node $i$ at propagation step $k$ for brevity.

For an arbitrary node $i$, each element $\mathbf{Z}_{i,m} (m\in\{1,... ,d\})$ of the output representation $\mathbf{Z}_{i} \in \mathbb{R}^{1\times d}$ learned by the decoupled GCN according to Eq. (\ref{eq:dgcn}) is calculated as:
\begin{equation}
	\mathbf{Z}_{i,m}= \sum_{k=0}^{K} \beta_{k}\mathbf{H}_{i,m}^{(k)}.
	\label{eq:n-dgcn}
\end{equation}

On the other hand, the output $\mathbf{X}_{i}\in \mathbb{R}^{(K+1)\times d}$ of \pname in the matrix form for node $i$ is described as:
\begin{equation}
   \mathbf{X}_{i} = \begin{bmatrix}
	\mathbf{H}_{i,0}^{(0)} & \mathbf{H}_{i,1}^{(0)} & \cdots & \mathbf{H}_{i,d}^{(0)} \\
	\mathbf{H}_{i,0}^{(1)} & \mathbf{H}_{i,1}^{(1)} & \cdots & \mathbf{H}_{i,d}^{(1)} \\
	\vdots&\vdots& \ddots&\vdots\\
	\mathbf{H}_{i,0}^{(K)} & \mathbf{H}_{i,1}^{(K)} & \cdots & \mathbf{H}_{i,d}^{(K)} \\
	\end{bmatrix}.
	\label{eq:x-n}
\end{equation}

Suppose we have the following attention matrix $\mathbf{S}\in \mathbb{R}^{(K+1)\times (K+1)}$:
\begin{equation}
   \mathbf{S} = \begin{bmatrix}
	0 & 0 & \cdots & 0 \\
	0 & 0 & \cdots & 0 \\
	\vdots&\vdots& \ddots&\vdots\\
	\beta_{0} & \beta_{1} & \cdots & \beta_{K} \\
	\end{bmatrix}.
	\label{eq:x-s}
\end{equation}

Following Eq. (\ref{eq:att-out}), the output matrix $\mathbf{T} \in \mathbb{R}^{(K+1)\times d}$ learned by the self-attention mechanism can be described as:
\begin{equation}
   \mathbf{T} = \mathbf{S} \mathbf{X}_{i} = \begin{bmatrix}
	0 & 0 & \cdots & 0 \\
	0 & 0 & \cdots & 0 \\
	\vdots&\vdots& \ddots&\vdots\\
	\gamma_{0} & \gamma_{1} & \cdots & \gamma_{d} \\
	\end{bmatrix},
	\label{eq:x-s}
\end{equation}
where $\gamma_{m} = \sum_{k=0}^{K} \beta_{k}\mathbf{H}_{i,m}^{(k)} (m\in\{1,... ,d\})$.

Further, we can obtain each element $\mathbf{T}_{m}^{final} (m\in\{1,... ,d\})$ of the final representation $\mathbf{T}^{final} \in \mathbb{R}^{1\times d}$ of node $i$ by using a summation readout function:
\begin{equation}
    \mathbf{T}_{m}^{final} = \sum_{k=0}^{K} \mathbf{T}_{k,m}=(0+0+\cdots+\gamma_{m})=\sum_{k=0}^{K} \beta_{k}\mathbf{H}_{i,m}^{(k)}=\mathbf{Z}_{i,m}.
	\label{eq:x-out}
\end{equation}

Finally, we can obtain \textbf{Fact 1}.
\hfill$\square$

\textbf{Fact 1} indicates that the decoupled GCN, an advanced category of GNN, only captures partial information of the multi-hop neighborhoods through the incomplete attention matrix.
Moreover, the fixed attention coefficients of $\beta_{k}$ ($k\in\{0,...,K\}$ ) for all nodes also limit the model to learn the node representations adaptively from their individual neighborhood information.

In contrast, our proposed \name first utilizes the self-attention mechanism to learn the representations of different-hop neighborhoods based on their semantic correlation.
Then, \name develops an attention-based readout function to adaptively learn the node representations from their neighborhood information, which helps the model learn more informative node representations.

%% file: 5DA.tex
Benefited from the proposed \pname that enables us to augment the graph data from the perspective of neighborhood information, in this section, we propose a novel graph data augmentation method called \DAname (\daname), which augments the sequences $\mathbf{X}_G$ obtained by \pname through two parts, namely \GNA (\gna) and \LNA (\lna), from the perspectives of global mixing and local destruction, respectively.

\subsection{\GNA}
Inspired by Mixup~\cite{mixup}, \GNA (\gna) aims to generate new training examples by mixing the information of pairwise nodes of the same training batch.

Specifically, we first decide whether to apply \gna with probability $p_{aug}$, which is a fixed hyper-parameter. Then we randomly combine two sequences of different nodes in a mini-batch, $\mathcal{S}^{i}_{v}$ and $\mathcal{S}^{j}_{v}$, to generate a new global augmentation sequence and its corresponding interpolating label $\tilde{\mathbf{Y}}_{v}$. The \gna sample can be described as: 
\begin{equation}
\begin{aligned}
        \tilde{{\mathcal{S}}}^{glo}_{v} &= \lambda\mathcal{S}^{i}_{v} + (1-\lambda)\mathcal{S}^{j}_{v},\\
        \tilde{\mathbf{Y}}_{v} &= \lambda{\mathbf{Y}_{i}} + (1-\lambda){\mathbf{Y}_{j}},
	\label{eq:gna}
\end{aligned}
\end{equation}
where we follow the setting in~\cite{mixup} and sample $\lambda$ from the beta distribution $Beta(\alpha,\beta)$, where $\alpha$ and $\beta$ control the shape of the beta distribution.

\subsection{\LNA}
The goal of \LNA (\lna) is to generate augmentation data examples by randomly masking a portion of the sequence obtained by \pname for each node, which could be regarded as local destruction to the original neighborhood information.

Specifically, for the $k$-hop neighborhood representation of node $v$, we randomly select some neighborhood representations to be masked. 
The corresponding augmented example $\tilde{\mathcal{S}}^{loc}_{v}$ is defined as:
\begin{equation}
        \tilde{\mathcal{S}}^{loc}_{v} = \mathbf{M} \odot (\mathbf{x}^0_v, \mathbf{x}^1_v,...,\mathbf{x}^K_v ),
	\label{eq:lna}
\end{equation}
where $\mathbf{M} \in {\{\mathbf{0},\mathbf{1}\}}^{d \times (K+1)}$ denotes a randomly generated binary mask that controls the area to mask. The column vectors in $\mathbf{M}$ are $d$-dimensional vectors filled with all zeros or all ones. Operator $\odot$ represents element-wise multiplication. In addition, we use a hyper-parameter $\tau$ to control the mask ratio. And we set the number of column vectors filled with zeros to $\lfloor (K+1) \times \tau \rfloor$ while ensuring that at least one column vector is filled with all zeros.

\subsection{\daname for Data Augmentation}
\label{use dataaug}
In the training phase, our \daname is adopted to generate new training samples by combining the output of two main modules, \gna and \lna.
In practice, each mini-batch will randomly determine whether to perform the \daname operator with probability $p_{aug}$ and returns the new sequences of all nodes in the mini-batch with the corresponding labels after augmentation. The resulting augmented data is then used to train the subsequent network.

The overall process is shown in Algorithm~\ref{alg:DA}.
It's worth noting that the core modules, \gna and \lna, could be applied independently to achieve neighborhood data augmentation. 
We conduct experiments to analyze the contributions of each module to the model performance in Section \ref{Ablation Study For DAname}.

\begin{algorithm}[t]  
	\caption{The \DAname Algorithm}  
	\label{alg:DA}  
	\begin{algorithmic}[1]  
		\Require
		Sequences of all nodes in a mini-batch $\mathcal{S}^{batch}$; Label matrix $\mathbf{Y}$; Mask ratio $\tau$; Probability $p_{aug}$; Shape parameters $\alpha$, $\beta$ of the beta distribution
		\Ensure  
		Augmented sequences of all nodes in a batch $\tilde{\mathcal{S}}^{batch}$; Augmented label matrix $\tilde{\mathbf{Y}}$
	    \If{$random(0,1)>p_{aug}$}
                \State $\tilde{\mathcal{S}}^{batch}=\mathcal{S}^{batch};\tilde{\mathbf{Y}}=\mathbf{Y};$
            \Else
                \State
                $\lambda=Beta(\alpha,\beta)$;
                
                \For{$\mathcal{S}^{i}_{v}$ in $\mathcal{S}^{batch}$}
                \State
                $\mathcal{S}^{j}_{v}$ = random\_sample($\mathcal{S}^{batch}$);
                \State
                Calculate $\tilde{{\mathcal{S}}}^{glo}_{v}$, $\tilde{\mathbf{Y}}$ by Eq. (\ref{eq:gna}) using $\mathcal{S}^{i}_{v}$, $\mathcal{S}^{j}_{v}$, $\lambda$, $\mathbf{Y}$; 
                \State
                Calculate $\tilde{\mathcal{S}}^{loc}_{v}$ by Eq. (\ref{eq:lna}) using $\tilde{{\mathcal{S}}}^{glo}_{v}$, $\tau$; 
                \EndFor
                \State
                $\tilde{\mathcal{S}}^{batch}=\mathcal{S}^{batch};$
                
            \EndIf
        \\
        \Return $\tilde{\mathcal{S}}^{batch}$, $\tilde{\mathbf{Y}}$;
	\end{algorithmic}  
\end{algorithm}

%% file: 6Exp.tex
 
In this section, we conduct extensive experiments to validate the effectiveness of \name, and the extended version of \name via \daname called \tname.
We first introduce the experimental setup.
Then we report the performances of \name and \tname against representative baselines on small-scale and large-scale real-world datasets. 
Finally, we provide the parameter and ablation studies to understand our proposed methods deeply.

\subsection{Experimental Setup}
Here we briefly introduce the datasets, baselines, and implementation details in our experiments. 

\begin{table}[t]
    \centering
	\caption{Statistics on datasets.}
	\label{tab:dataset}

	\begin{tabular}{lrrrrc}
		\toprule
		Dataset & \# Nodes & \# Edges & \# Features & \# Classes \\
		\midrule
		Pubmed & 19,717 & 44,324 & 500 & 3\\
		CoraFull & 19,793 & 126,842 & 8,710 & 70\\
		Computer & 13,752 & 491,722 & 767 & 10\\
		Photo & 7,650 & 238,163 & 745 & 8\\
		CS & 18,333 & 163,788 & 6,805 & 15\\
		Physics & 34,493 & 495,924 & 8,415 & 5\\
		AMiner-CS & 593,486 & 6,217,004 & 100 & 18\\
		Reddit & 232,965 & 11,606,919 & 602 & 41\\
		Amazon2M & 2,449,029 & 61,859,140 & 100 & 47\\
		\bottomrule
	\end{tabular}
\end{table}

\textbf{Datasets}. 
We conduct experiments on nine widely used datasets of various scales, including six small-scale datasets and three relatively large-scale datasets. 
For small-scale datasets, we adopt Pubmed, CoraFull, Computer, Photo, CS and Physics from the Deep Graph Library (DGL). 
We apply 60\%/20\%/20\% train/val/test random splits for small-scale datasets.
For large-scale datasets, we adopt AMiner-CS, Reddit and Amazon2M from~\cite{Grand+}.
The splits of large-scale datasets follow the settings of ~\cite{Grand+}.
Statistics of the datasets are reported in Table~\ref{tab:dataset}.

\textbf{Baselines}. 
We compare \name with 12 advanced baselines, including: 
(\uppercase\expandafter{\romannumeral1}) four full-batch GNNs:~GCN~\cite{GCN}, GAT~\cite{GAT}, APPNP~\cite{appnp} and GPRGNN~\cite{gprgnn}; 
(\uppercase\expandafter{\romannumeral2}) three scalable GNNs:~GraphSAINT~\cite{GraphSAINT}, PPRGo~\cite{PPRGo} and GRAND+~\cite{Grand+}; 
(\uppercase\expandafter{\romannumeral3}) five graph Transformers\footnote{Another recent graph Transformer, SAT~\cite{sat}, is not considered as it reports OOM even in our small-scale graphs.}:~GT~\cite{GT}, SAN~\cite{SAN}, Graphormer~\cite{Graphormer}, GraphGPS~\cite{gps} and NodeFormer~\cite{nodeformer}.

\textbf{Implementation details}.
Referring to the recommended settings in the official implementations, we perform hyper-parameter tuning for each baseline.  
For the model configuration of \name, we try the number of Transformer layers in $\{1,2,...,5\}$, the hidden dimension in $\{128, 256, 512$\}, and the propagation steps in $\{2,3,...,20\}$.
Parameters are optimized with the AdamW~\cite{adamw} optimizer, using a learning rate of in $\{1e-3,5e-3,1e-4\}$ and the weight decay of $\{1e-4,5e-4,1e-5\}$. 
We also search the dropout rate in $\{0.1,0.3,0.5\}$.
We follow the setting in~\cite{mixup} and set the shape parameters $\alpha$ and $\beta$ of beta distribution in \gna to $1.0$.
The batch size is set to 2000.
The training process is early stopped within 50 epochs.
For the hyper-parameters of \daname, we try the mask ratio $\tau$ in $\{0.25,0.5,0.75\}$, and the augmented probability $p_{aug}$ in $\{0.25,0.5,0.75,1.0\}$.
All experiments are conducted on a Linux server with 1 I9-9900k CPU, 1 RTX 2080TI GPU and 64G RAM. 

\begin{table*}[t]
    \centering
	\caption{
	Comparison of all models in terms of mean accuracy $\pm$ stdev (\%) on small-scale datasets. 
	The best results appear in \textbf{bold}. OOM indicates the out-of-memory error.
	}
	\label{tab:normal-data}
	\setlength\tabcolsep{1mm}{
	\scalebox{1.0}{
	\begin{tabular}{lcccccc}
		\toprule
		Method  &Pubmed &CoraFull &Computer &Photo &CS &Physics\\
		\midrule
		GCN &86.54 $\pm$ 0.12  & 61.76 $\pm$ 0.14 & 89.65 $\pm$ 0.52  & 92.70 $\pm$ 0.20 & 92.92 $\pm$ 0.12 & 96.18 $\pm$ 0.07\\
		GAT &86.32 $\pm$ 0.16  & 64.47 $\pm$ 0.18 & 90.78 $\pm$ 0.13  & 93.87 $\pm$ 0.11 & 93.61 $\pm$ 0.14 & 96.17 $\pm$ 0.08 \\
		APPNP &88.43 $\pm$ 0.15  & 65.16 $\pm$ 0.28 & 90.18 $\pm$ 0.17  & 94.32 $\pm$ 0.14 & 94.49 $\pm$ 0.07 & 96.54 $\pm$ 0.07\\
		GPRGNN &89.34 $\pm$ 0.25  & 67.12 $\pm$ 0.31 & 89.32 $\pm$ 0.29  & 94.49 $\pm$ 0.14 & 95.13 $\pm$ 0.09 & 96.85 $\pm$ 0.08\\
		\midrule
		GraphSAINT & 88.96 $\pm$ 0.16  & 67.85  $\pm$ 0.21  & 90.22 $\pm$ 0.15  & 91.72 $\pm$ 0.13  & 94.41 $\pm$ 0.09 & 96.43 $\pm$ 0.05  \\
		PPRGo & 87.38 $\pm$ 0.11   & 63.54 $\pm$ 0.25 & 88.69 $\pm$ 0.21  & 93.61 $\pm$ 0.12  & 92.52 $\pm$ 0.15 & 95.51 $\pm$ 0.08  \\
		GRAND+ & 88.64 $\pm$ 0.09   & 71.37 $\pm$ 0.11 &  88.74 $\pm$ 0.11   & 94.75 $\pm$ 0.12  & 93.92 $\pm$ 0.08 & 96.47 $\pm$ 0.04   \\
		\midrule
		GT &88.79 $\pm$ 0.12  & 61.05 $\pm$ 0.38 & 91.18 $\pm$ 0.17  & 94.74 $\pm$ 0.13 & 94.64 $\pm$ 0.13 & 97.05 $\pm$ 0.05 \\
		Graphormer & OOM   &  OOM  &  OOM   & 92.74 $\pm$ 0.14 &  OOM  &  OOM \\
		SAN & 88.22 $\pm$ 0.15   & 59.01 $\pm$ 0.34 & 89.83 $\pm$ 0.16  & 94.86 $\pm$ 0.10 & 94.51 $\pm$ 0.15 &  OOM \\
	    GraphGPS & 88.94 $\pm$ 0.16   & 55.76 $\pm$ 0.23 & OOM  & 95.06 $\pm$ 0.13 & 93.93 $\pm$ 0.12 &  OOM  \\
	    NodeFormer & 89.24 $\pm$ 0.14   & 61.82 $\pm$ 0.25 & 91.12 $\pm$ 0.19  & 95.27 $\pm$ 0.17 & 95.68 $\pm$ 0.08 &  97.19 $\pm$ 0.04  \\
		\midrule
		\name &{89.70 $\pm$ 0.19}  & {71.51 $\pm$ 0.13} &  {91.22 $\pm$ 0.14}  & {95.49 $\pm$ 0.11} & {95.75 $\pm$ 0.09} & \textbf{97.34 $\pm$ 0.03} \\
            \tname & \textbf{90.38 $\pm$ 0.20}&\textbf{72.16 $\pm$ 0.29} &\textbf{91.95 $\pm$ 0.09} & \textbf{96.61 $\pm$ 0.21}& \textbf{96.06 $\pm$ 0.10}&{97.34 $\pm$ 0.09} \\
		\bottomrule
	\end{tabular}}
	}
\end{table*}

\subsection{Comparison on Small-scale Datasets}
We conduct 10 trials with random seeds for
each model and take the mean accuracy and standard deviation for comparison on small-scale datasets, and the results are reported in Table~\ref{tab:normal-data}.  
From the experimental results, we can observe that \name outperforms the baselines consistently on all these datasets.
For the superiority over GNN-based methods, it is because 
\name utilizes \pname and the Transformer model to capture the semantic relevance of different hop neighbors overlooked in most GNNs, especially compared to two decoupled GCNs, APPNP and GPRGNN.
Besides, the performance of \name also surpasses graph Transformer-based methods, indicating that leveraging the local information is beneficial for node classification. 
In particular, \name outperforms GT and SAN, which also introduce the eigenvectors of Laplacian matrix as the structural encoding into Transformers for learning the node representations, demonstrating the superiority of our proposed \name.
Moreover, We observe that Graphormer, SAN, and GraphGPS suffer from the out-of-memory error even in some small graphs, further demonstrating the necessity of designing a scalable graph Transformer for large-scale graphs. 
Finally, it is noteworthy that \tname surpasses all models and leads to state-of-the-art results, which shows the performance of \name has been further improved upon incorporating \daname, indicating that \daname effectively augments the input data from small-scale datasets.

\begin{table}[t]
    \centering
	\caption{
	Comparison of all models in terms of mean accuracy $\pm$ stdev (\%) on large-scale datasets. 
	The best results appear in \textbf{bold}.
	}
	\label{tab:large-data}
\scalebox{1.0}{
	\begin{tabular}{lccc}
		\toprule
		Method  &AMiner-CS &Reddit &Amazon2M\\
		\midrule
		PPRGo & 49.07 $\pm$ 0.19 & 90.38 $\pm$ 0.11  & 66.12 $\pm$ 0.59 \\
		GraphSAINT & 51.86  $\pm$ 0.21 &  92.35 $\pm$ 0.08   & 75.21 $\pm$ 0.15  \\
		GRAND+ & 54.67 $\pm$ 0.25  &92.81 $\pm$ 0.03  & 75.49 $\pm$ 0.11   \\
		\midrule
		\name  &{56.21 $\pm$ 0.42}  & {93.58 $\pm$ 0.05} & {77.43 $\pm$ 0.24}\\
            \tname   &\textbf{57.02 $\pm$ 0.38} & \textbf{93.74 $\pm$ 0.06} &\textbf {77.98 $\pm$ 0.16}\\
		\bottomrule
	\end{tabular}
 	}
\end{table}

\subsection{Comparison on Large-scale Datasets}
To verify the scalability of \name and \tname, we continue the comparison on three large-scale datasets. 
For the baselines, we only compare with three scalable GNNs, as existing graph Transformers can not directly work on such large-scale datasets due to their high training cost. 
The results are summarized in Table \ref{tab:large-data}.
One can see that \name consistently outperforms the scalable GNNs on all datasets, indicating that \name can better preserve the local information of nodes and is capable of handling the node classification task in large graphs.
Furthermore, \tname boosts the performance on all the three datasets, showing that \tname can still effectively perform the node classification task on large-scale datasets.

\begin{table*}[t]
	\caption{
    The accuracy (\%) with or without structural encoding.
	}
 \label{tab:ab-str}
 \centering
 \renewcommand\arraystretch{1.3}
 \setlength\tabcolsep{1.5mm}{
		\footnotesize
  \scalebox{1.0}{
\begin{tabular}{clccccccccc}
\toprule
                                                  &         & Pumbed & Corafull & CS    & Computer & Photo & Physics & Aminer-CS & Reddit & Amazon2M \\ \hline
\multicolumn{1}{c|}{\multirow{3}{*}{NAGphormer}}  & W/O-SE  & 89.06  & 70.42    & 95.52 & 90.44    & 95.02 & 97.10   & 55.64     & 93.47  & 76.98    \\
\multicolumn{1}{c|}{}                             & With-SE & 89.70  & 71.51    & 95.75 & 91.22    & 95.49 & 97.34   & 56.21     & 93.58  & 77.43    \\ \cline{2-11} 
\multicolumn{1}{c|}{}                             & Gain    & +0.64  & +1.09    & +0.23 & +0.78    & +0.47 & +0.24   & +0.57     & +0.11  & +0.45    \\ \hline
\multicolumn{1}{c|}{\multirow{3}{*}{NAGphormer+}} & W/O-SE  & 90.09  & 71.81    & 95.90 & 91.60    & 96.20 & 97.27   & 56.14     & 93.53  & 77.78    \\
\multicolumn{1}{c|}{}                             & With-SE & 90.38  & 72.16    & 96.06 & 91.95    & 96.61 & 97.34   & 57.02     & 93.74  & 77.98    \\ \cline{2-11} 
\multicolumn{1}{c|}{}                             & Gain    & +0.29  & +0.35    & +0.16 & +0.35    & +0.41 & +0.07   & +0.88      & +0.21  & +0.20     \\ \toprule
\end{tabular}}}
\end{table*}

\begin{figure}[t]
    \centering
	\includegraphics[width=3.5in]{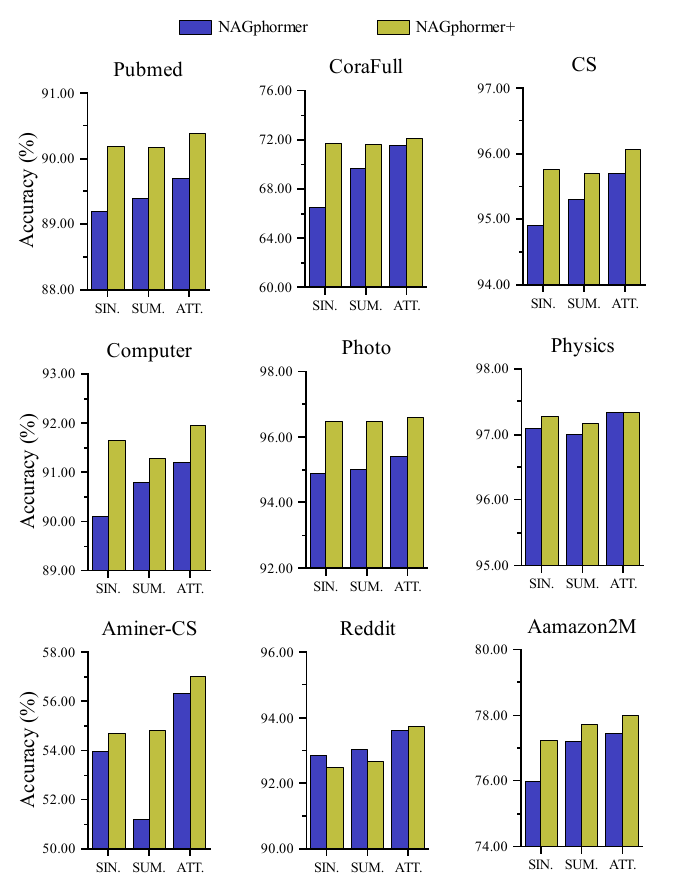}
	\caption{The performance of \name with different readout functions.}
	\label{fig:rd}
\end{figure}

\subsection{Ablation Study}
To analyze the effectiveness of structural encoding and attention-based readout function, we perform a series of ablation studies on all datasets. 

\textbf{Structural encoding}.
We compare our proposed \name and \tname to its variant without the structural encoding module to measure the gain of structural encoding. 
The results are summarized in Table~\ref{tab:ab-str}.
We can observe that the gains of adding structural encoding vary in different datasets, since different graphs exhibit different topology structure. 
Therefore, the gain of structural encoding is sensitive to the structure of graphs. 
These results also indicate that introducing the structural encoding can improve the model performance for the node classification task.

\textbf{Attention-based readout function}.
We conduct a comparative experiment between the proposed attention-based readout function $\mathrm{ATT.}$ in Eq. (\ref{eq:att_readout}) with previous readout functions, \ie, $\mathrm{SIN.}$ and $\mathrm{SUM.}$. 
The function of $\mathrm{SIN.}$ utilizes the corresponding representation of the node itself learned by the Transformer layer as the final output to predict labels.  
And $\mathrm{SUM.}$ can be regarded as aggregating all information of different hops equally. 
We evaluate the performance of \name and \tname with different readout functions on all benchmark datasets.
From Figure~\ref{fig:rd}, we observe that $\mathrm{ATT.}$ consistently outperforms other readout functions on all benchmark datasets, 
indicating that aggregating information from different neighborhoods adaptively is beneficial to learn more expressive node representations, 
further improving the model performance on node classification.

\begin{figure}[t]
    \centering
	\includegraphics[width=3.5in]{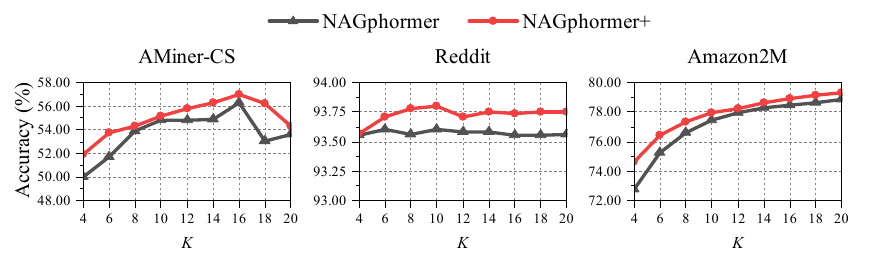}
	\caption{On the number of propagation steps $K$.}
	\label{fig:exp3_k}
\end{figure}

\begin{figure}[t]
    \centering
	\includegraphics[width=3.5in]{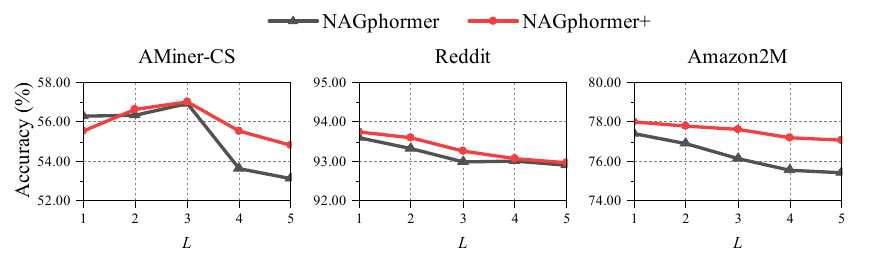}
	\caption{On the number of propagation steps $L$.}
	\label{fig:exp3_l}
\end{figure}

\subsection{Parameter Study}

To further evaluate the performance of \name and \tname, we study the influence of two key parameters: the number of propagation steps $K$ and the number of Transformer layers $L$. Specifically, we perform experiments on AMiner-CS, Reddit and Amazon2M by setting different values of $K$ and $L$, respectively.

\textbf{On parameter $K$}.
We fix $L=1$ and vary the number of propagation steps $K$ in $\{4,6,\cdots, 20\}$.
Figure \ref{fig:exp3_k} reports the model performance. 
We can observe that the values of $K$ are different for each dataset to achieve the best performance since different networks exhibit different neighborhood structures.
Besides, we can also observe that the model performance does not decline significantly even if $K$ is relatively large to 20. 
For instance, the performance on Reddit dataset changes slightly ($<0.1\%$) with the increment of $K$, which indicates that learning the node representations from information of multi-hop neighborhoods via the self-attention mechanism and attention-based readout function can alleviate the impact of over-smoothing and over-squashing problems.
In addition, the model performance changes differently on three datasets with the increment of $K$.
The reason may be that these datasets are different types of networks and have diverse properties.
This observation also indicates that neighborhood information on different types of networks has different effects on the model performance.
In practice, we set $K=16$ for AMiner-CS, and set $K=10$ for others since the large propagation step will bring the high time cost of \pname on Amanzon2M.

\textbf{On parameter $L$}.
We fix the best value of $K$ and vary $L$ from 1 to 5 on each dataset. The results are shown in Figure \ref{fig:exp3_l}. 
Generally speaking, a smaller $L$ can achieve a high accuracy while a larger $L$ degrades the performance of \name and \tname. 
Such a result can attribute to the fact that a larger $L$ is more likely to cause over-fitting.
we set $L=3$ for AMiner-CS, and set $L=1$ for other datasets.

It is worth noting that the variation trend of \tname under different parameters is essentially consistent with that of \name, indicating that the \daname we designed is highly suitable for \name.

\begin{figure}[t]
    \centering
	\includegraphics[width=3.5in]{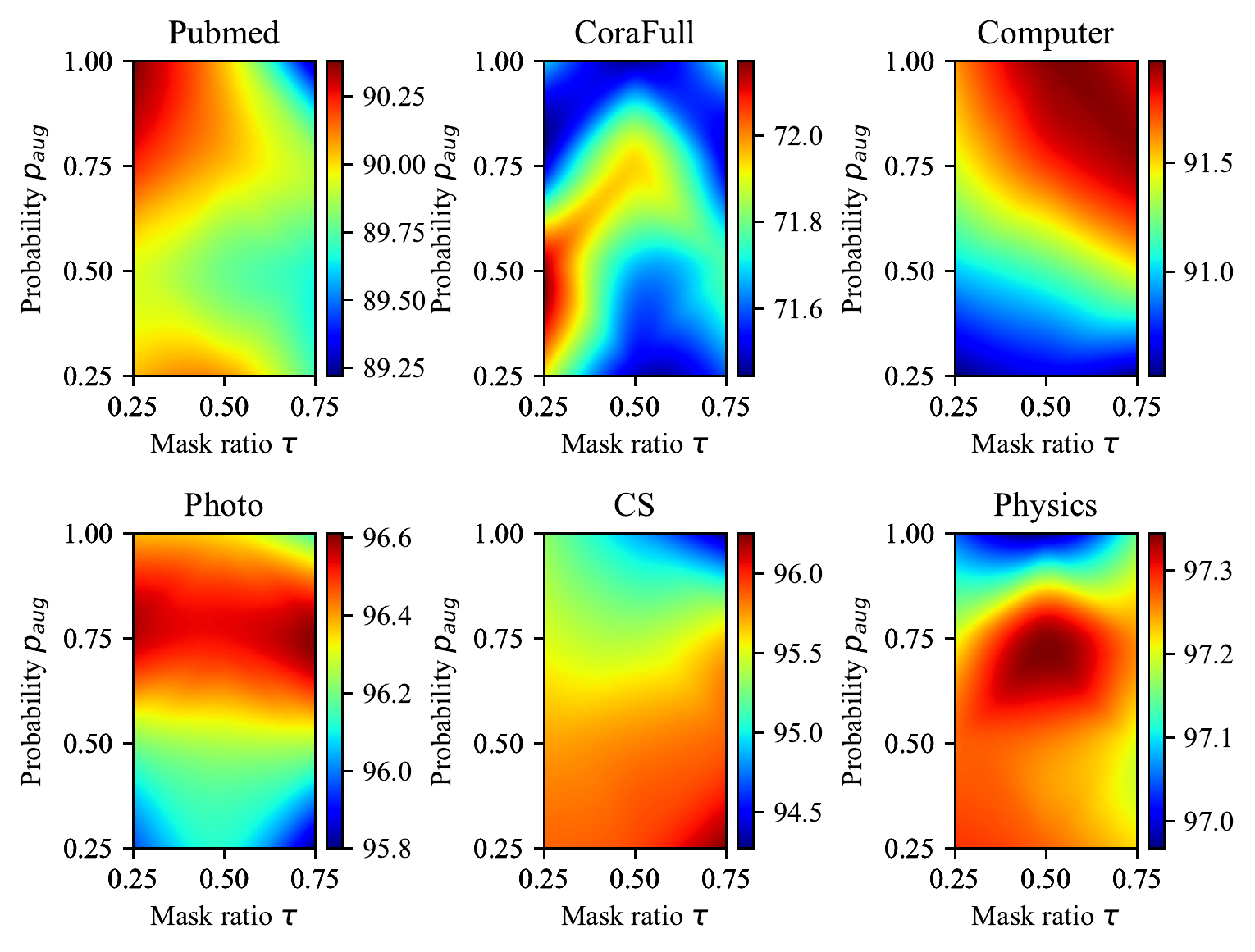}
	\caption{Accuracy (\%) with different hyper-parameters of \daname. }
	\label{fig:DA_ablation}
\end{figure}

\begin{table*}[t]
    \centering
    \caption{The accuracy (\%) of \name with different augmentation methods. The best results appear in \textbf{bold}.}
    \label{tab:DA_result}
     \renewcommand\arraystretch{1.3}
    \setlength\tabcolsep{1.5mm}{
		\footnotesize
	\scalebox{1.1}{
\begin{tabular}{lccccccccc}
\toprule
  & Pubmed & CoraFull & Computer & Photo & CS & Physics & Aminer-CS & Reddit & Amazon2M \\ \hline
\name &  89.70   &  71.51  &  91.22 & 95.49 & 95.75 & \textbf{97.34} & 56.21 & 93.58  & 77.43  \\ 
+\lna &  89.89 &  71.83   &  91.88  &  96.36 & 95.89 &  97.23  &  55.88 & 93.26 & 77.37  \\
+\gna &  90.31 & 72.11 &  90.99  & 96.36 & 95.99 & 97.25 & 56.56  & 93.55 & 77.60  \\ \hline
+\daname & \textbf{90.38} & \textbf{72.16} & \textbf{91.95} &\textbf{96.61} &\textbf{96.06} &97.34 &\textbf{57.02} &\textbf{93.74} & \textbf{77.98}\\ \toprule
\end{tabular}}}
\end{table*}

\begin{table*}[t]
\centering
\caption{The training cost on large-scale graphs in terms of GPU memory (MB) and running time (s).}
\renewcommand\arraystretch{1.3}
\footnotesize
\label{tab:cost}
\begin{tabular}{lcrcrcr}
\toprule
&\multicolumn{2}{c}{Aminer-CS}&\multicolumn{2}{c}{Reddit}&\multicolumn{2}{c}{Amazon2M} \\
& Memory (MB)  & Time (s)  & Memory (MB) & Time (s) & Memory (MB) & Time (s) \\ \hline
GraphSAINT & 1,641& 23.67 & 2,565 & 43.15& 5,317& 334.08  \\ 
PPRGo& 1,075&14.21& 1,093 &35.73 &1,097& 152.62 \\ 
GRAND+&1,091&21.41&1,213 & 197.97 &1,123& 207.85 \\ \hline
NAGphormer & 1,827 &19.87& 1,925&20.72& 2,035  & 58.66               \\
NAGphormer+ & 2,518 &26.92& 2,706&46.80& 2,290  & 65.22               \\\toprule
& \multicolumn{1}{l}{} & \multicolumn{1}{l}{} & \multicolumn{1}{l}{} & \multicolumn{1}{l}{} & \multicolumn{1}{l}{} & \multicolumn{1}{l}{}
\end{tabular}
\end{table*}

\subsection{Analysis of \daname}
\label{Ablation Study For DAname}
We further conduct additional experiments to deeply analyze our proposed \daname.
As mentioned in Section \ref{use dataaug}, the two core modules of \daname, \gna and \lna, could be applied independently for augmenting the input data of \name.
Hence, we first conduct experiments to evaluate the performance of \name via only \gna or \lna.
The results are reported in Table~\ref{tab:DA_result}.

As shown in Table~\ref{tab:DA_result}, \lna can significantly improve \name's performance on both Computer and Photo datasets. 
And \gna can dramatically improve \name's performance on various datasets, including Pubmed, CoraFull, and Photo. 
The results indicate that \gna has a superior effect compared to \lna, which could be attributed to its use of mixed samples generated from multiple nodes’ data to capture a wider range of neighborhood information.
Furthermore, the two methods continue to improve the performance on most datasets when combined, demonstrating that the two methods can complement each other.

Then, we study the influence of two key hyper-parameters, the mask ratio $\tau$ for \lna and the probability $p_{aug}$, on the model performance.
For simplicity, we set $K$ to the value at which the searched parameter yields the best performance for \name. 
We interpolate the test accuracy after the grid search for hyper-parameters. 
As shown in Figure \ref{fig:DA_ablation}, applying \daname on different datasets to achieve the best performance requires different hyper-parameters. 
Generally speaking, a larger value of $p_{aug}$ tends to result in a more effective combination of \gna and \lna.

In a word, it is clear that we can easily implement \daname leveraging the benefits of \pname, which can transform irregular and non-Euclidean data into structural data. 
Experiments demonstrate that our \daname can improve the model performance, further highlighting the innovative nature of \name.

\subsection{Efficiency Study}\label{efficiency}
In this subsection, we validate the efficiency of \name and \tname on large-scale graphs.
Specifically, we compare the training cost in terms of running time (s) and GPU memory (MB) of \name, \tname and three scalable GNNs, PPRGo, GraphSAINT and GRAND+.
For scalable GNNs, we adopt the official implements on Github.
However, all methods contain diverse pre-processing steps built on different programming language frameworks, such as approximate matrix-calculation based on C++ framework in GRAND+.
To ensure a fair comparison, we report the running time cost including the training stage and inference stage since these stages of all models are based on Pytorch framework.
The results are summarized in Table \ref{tab:cost}.

From the results, we can observe that \name shows high efficiency when dealing with large graphs. For instance, on Amazon2M which contains two million nodes and 60 million edges, \name achieves almost $3\times$ acceleration compared with the second fastest model PPRGo. The reason is that the time complexity of \name mainly depends on the number of nodes and 
is not related to the number of edges, while the time consumption of other methods is related to the number of both edges and nodes since these methods involve the propagation operation during the training and inference stages.
And the increase in time required for \tname compared to \name may be attributed to \daname providing more challenging data, thereby causing the network to spend more time on learning. But the additional time required is acceptable.
As for the GPU memory cost, since \name utilizes the mini-batch training, the GPU memory cost is determined by the batch size. 
Hence, the GPU memory cost of \name and \tname is affordable by choosing a proper batch size even on large-scale graphs.